\documentclass{AISB2008}
\usepackage{times}
\usepackage{graphicx}
\usepackage{latexsym}
\usepackage{todonotes}
\usepackage{balance}

\begin{document}

\title{Physical extracurricular activities in educational child-robot interaction}

\author{Daniel Davison\institute{Dept. of Electrical Engineering, Mathematics and Computer Science, HMI, University of Twente, Enschede, The Netherlands. Corresponding author e-mail: d.p.davison@utwente.nl} \and Louisa Schindler\textsuperscript{\textmd1} \and Dennis Reidsma\textsuperscript{\textmd1}}

\maketitle

\begin{abstract}
In an exploratory study on educational child-robot interaction we investigate the effect of alternating a learning activity with an additional shared activity. Our aim is to enhance and enrich the relationship between child and robot by introducing ``physical extracurricular activities''. This enriched relationship might ultimately influence the way the child and robot interact with the learning material. We use qualitative measurement techniques to evaluate the effect of the additional activity on the child-robot relationship. We also explore how these metrics can be integrated in a highly exploratory cumulative score for the relationship between child and robot. This cumulative score suggests a difference in the overall child-robot relationship between children who engage in a physical extracurricular activity with the robot, and children who only engage in the learning activity with the robot.
\end{abstract}

\section{INTRODUCTION}
\begin{figure}
\centerline{\includegraphics[width=3.5in]{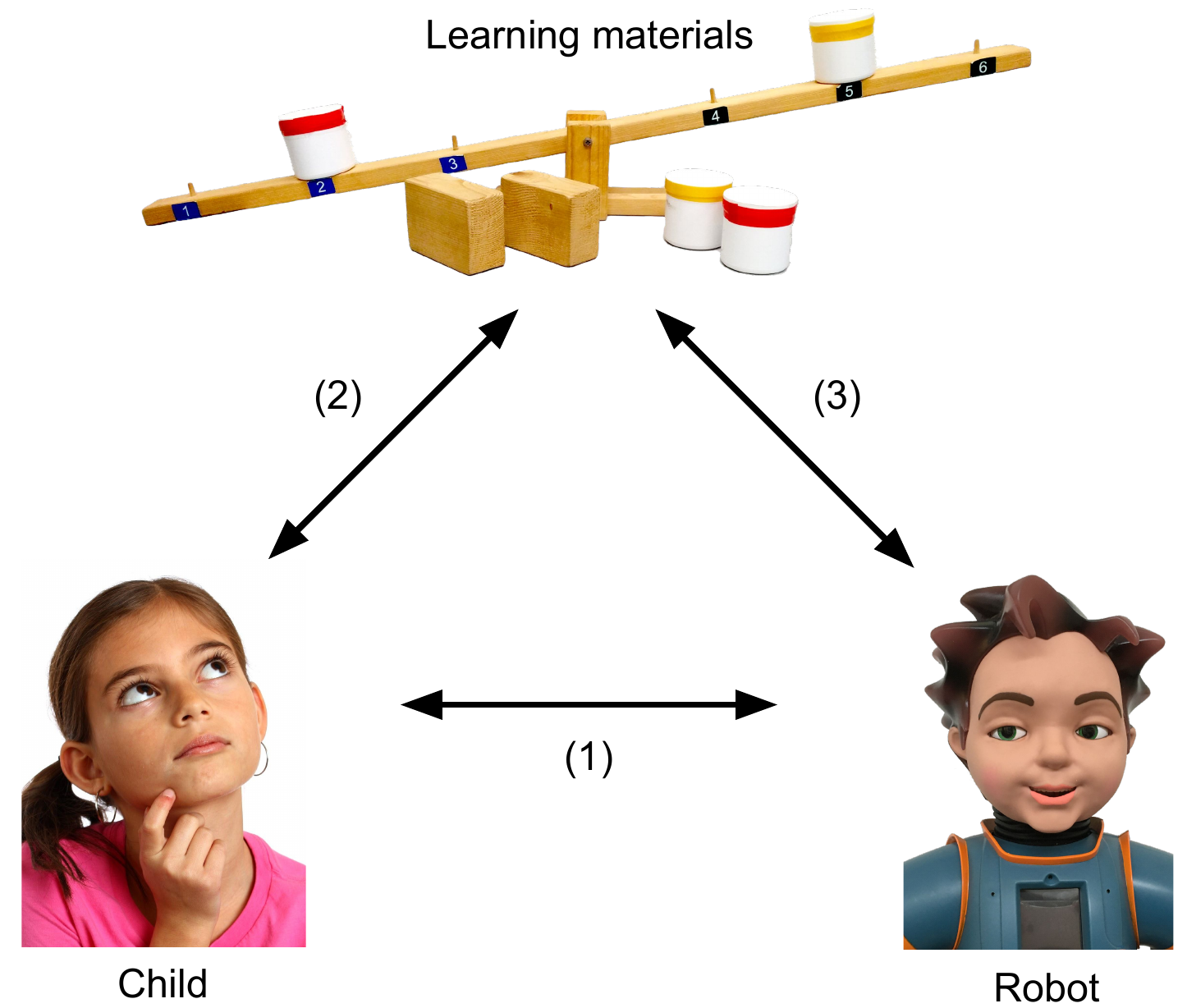}}
\caption{Schematic illustration of the triadic interaction between child, robot and learning materials. This triadic interaction consists of three distinct dyadic interactions: 1) child and robot; 2) child and learning materials; and 3) robot and learning materials. Each of these dyadic interactions is expected to influence the other two dyadic interactions to greater or lesser extent.} \label{fig:interaction_triangle}
\end{figure}

This paper discusses an exploratory study in which we investigate the relationship between a child and a robot working together to solve a learning task. In order to support children in their learning process, the relationship between the learner and the teacher or peer is crucial \cite{Rogoff1998, Vygotsky1978}. Within this context, a child, a robot and the learning materials are engaged in a triadic interaction, as illustrated in figure~\ref{fig:interaction_triangle}. The child interacts with the learning materials, together with the robot, in a collaborative learning setting. Interactions between child and robot and the relationship they form can influence how the child performs in the learning task and ultimately how the child learns \cite{Charisi2015}. The triadic interaction consists of three distinct dyadic interactions, which influence each other to greater or lesser extent: 1) interaction between child and robot; 2) interaction between child and learning materials; and 3) interaction between robot and learning materials.

Research in specific zones of this triadic interaction between child, robot and learning materials often focus specifically on one of the dyadic interactions, or on influences between these three dyadic interactions. Typical examples of research on the dyadic interaction between \emph{(1) child and robot} are those of Kahn et al. \cite{Kahn2012} and Kanda et al. \cite{Kanda2012}, who investigate children's perceptions of the robot and relationships with the robot, in an educational context. Typical examples of studies that show how interactions between \emph{(1) child and robot} influence the interactions between \emph{(2) child and learning materials} are: Kory and Breazeal \cite{Kory2014}, who investigate how matching a robot's competence level to that of the child influences the child's learning; and Chandra et al. \cite{Chandra2015}, who show that children feel more responsible in the learning task when working with a robot facilitator.

We investigate the dyadic interaction between \emph{(1) child and robot}, in which the child and peer-like robot engage together in a physical extracurricular activity, in an educational context. Since the robot is not necessarily presented to the child as a teacher, it could enrich the learning through implicit interaction in contrast to explicit teaching and assessment. However, the study presented here does not focus on measuring these potential effects on learning.

From Vygotsky's theories on child development and learning, we know that social scaffolding can be an important method for a child to transcend from his level of actual development to his level of potential development \cite{Vygotsky1978}. Previous work has shown indications that children working together with a social robot interact differently with learning materials, when compared to working with a less social tablet \cite{Wijnen2015}. We expect that a robot's social and relational features will impact a child's perception of the robot, and will influence their collaborative long-term interactions with learning materials. For such long-term interactions to take place, Belpaeme et al. \cite{Belpaeme2012} stress the importance of robot adaptability to a user's social needs. For example, Kanda et al. \cite{Kanda2007, Kanda2012} found that a friendly relationship between child and robot is one of the contributing factors for successful long-term use in a classroom setting. 

\begin{figure}
\centerline{\includegraphics[width=2in]{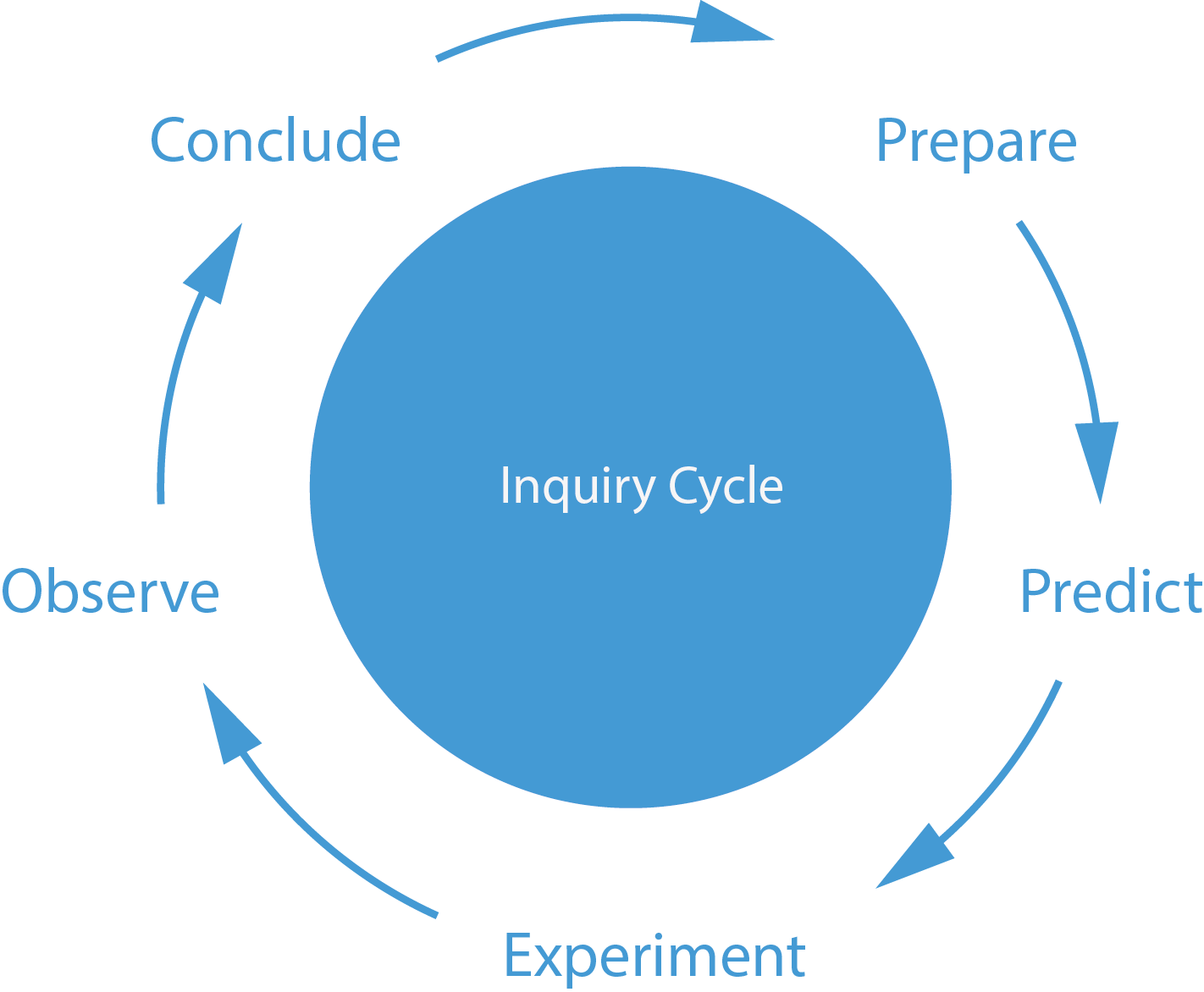}}
\caption{An adaptation of the inquiry cycle, simplified to match the skill level of primary school children.} \label{fig:inquiry_cycle}
\end{figure}

In this study we explore a possible method for enriching the child-robot relationship, through a shared extracurricular physical activity. This activity is performed in an educational context, and is introduced to the child as a short break from learning. The educational assignment is based on an inquiry learning model, in which a child discovers properties of the learning materials using a scientific approach. Hypothesis generation, experimentation, and evidence evaluation are often described as the core processes involved in scientific discovery learning \cite{Klahr1988, Klahr2000, Joolingen1997, Zimmerman2000}. A typical structured inquiry learning scenario involves multiple small assignments of increasing difficulty, in which the learner will go through a cycle of processes. Figure \ref{fig:inquiry_cycle} shows a adapted version of the inquiry cycle, which has been simplified to match the skill level of primary school children. Five distinct processes are included in this structured inquiry task: 1) prepare; 2) predict; 3) experiment; 4) observe; and 5) conclude.

For the specific inquiry task in this study, the learner uses a balance beam to discover the ``moment of force''. The balance beam acts as a weighing scale, with which the children can measure the relative weights of various available objects. By placing the objects at different offsets from the central pivot point, they discover that the moment of force acting on the balance is influenced by both the weight of the object and its distance from the pivot.

The non-educational activity is a short physical exercise with tasks such as: standing on one leg while raising the arms; standing on one leg while drawing the shape of an eight in the air; and closing the eyes or spelling the alphabet. In such tasks, physical and cognitive challenges are combined. We incorporated the cognitive challenges to make the difficulty level of the task higher (the combination also makes the physical activity harder to carry through). This leads to a challenging task for the child which nevertheless is very different from the educational task.

The robot used in this study is the Zeno R25, a humanoid robot developed by RoboKind. The Zeno has five degrees of freedom in his face, with which he can make basic facial expressions and emotions such as ``happy'', ``sad'' or ``surprised''. In addition, the robot is able to move his limbs and can shift his gaze using two degrees of freedom in his neck, combined with eye movements.

The goals for this study consist of two parts: Firstly, we explore if a physical extracurricular activity affects a child's perceived relationship with the robot. Secondly, we explore novel methods for conducting semi-structured interviews, which can be used to measure a change in this perceived relationship. Since written questionnaires are often not suitable for children of this age, we use a mixture of half-open interview questions, sociometric questions and small picture tasks. In the analysis we focus not only on the given answers, but take into account the child's reasoning behind these answers.

\section{METHOD}
We focus on learners between 6 and 11 years in age. It is the purpose of this study to research whether an additional joint activity (next to a learning activity) has positive influence on the relationship between robot and child. Therefore, a randomised controlled trial has been conducted between two groups of participants. The \emph{intervention group} performed two assignments with the learning material, then they continued with performing the intermediate physical exercise and subsequently they continued with two more assignments with the learning material. The \emph{control group} completed six learning task assignments together with the robot without performing the intermediate physical exercise. Consecutive learning tasks were of increasing difficulty, while ensuring that the difficulty of the first and last tasks was identical in both conditions.

To keep the remaining conditions as similar as possible, the overall time each child shared with the robot was approximately similar. The physical activity is therefore approximately as time consuming as two learning assignments. In addition, the robot's behaviour and movements are similar in both conditions. For instance, in both conditions the robot uses gaze, moves his head accordingly, and moves the body as natural as possible. Furthermore, the voice of the robot is equal in tone and emphasis, and the robot uses inviting forms of phrasing the sentences. For instance, the robot uses the phrase ``let's do (...)'' to trigger the next activity, and uses only positive phrases while supporting the child, such as ``well done'' or ``good job''. 

\subsection{Learning activities}
The learning activities are based on an inquiry learning cycle, during which the children go through several processes related to scientific discovery. They generally go through the following five processes for each learning assignment: 1) prepare the experiment; 2) predict the outcome (hypothesise); 3) perform the experiment; 4) observe the outcome; and 5) draw a conclusion. The adapted inquiry cycle is illustrated in figure \ref{fig:inquiry_cycle}. Consecutive assignments are generally of increasing difficulty, but all follow the same processes outlined by the inquiry cycle.

The specific inquiry activity used in this study is related to discovering the ``moment of force'': the children use a balance beam with a central pivot, to explore the effects of weight distribution on the forces acting on the balance. For instance, they discover that placing a heavy weight close to the pivot, will result in an equal force as placing a light weight far from the pivot.

\subsection{Physical extracurricular activity}
In addition to the learning activities, the children in the intervention group participate in a physical extracurricular activity with the robot. This physical activity was introduced by the robot as a break from studying. The robot invites the child to do a physical exercise together, such as ``Stand on one leg, and wave your arms, while reciting the alphabet'', or ``Close your eyes while moving your arms forward and standing on one leg''. The robot performs most of the demanded movements as well, however there are some restrictions. The robot can move the arms up and down, close the eyes, move the head, and stand on one leg. Some of the more complex movements are impossible for the robot, such as ``Stand on one leg and draw a `figure 8' in the air with the raised foot''. Therefore, the robot not only performs the actions, but also verbally explains what the child should do. 

\subsection{Measurement methods}
The goal of this study is to evaluate the relationship between child and robot. There are several methods that can be used in order to learn about the connection between the robot and the child. Most of these measures can be seen as tools with which to conduct a semi-structured interview with children. Examples of such tools can be found in the Fun Toolkit, constructed by Read et al. \cite{Read2002, Read2006}. Several of the methods discussed below are inspired by the Fun Toolkit, and have been adapted to match the target group's age. 

The measurements can be subdivided in three main categories, all of which are collected in the form of a semi-structured interview:

\begin{enumerate}
\item Pictorial task
\item Social distance task
\item Sociometric questions
\end{enumerate}

Firstly, an assignment is used in which children should describe their thoughts by means of a pictorial task. This approach supports children to describe situations or imperceptible concepts like relationships more precisely \cite{Harter1984}. First, the child draws himself or herself in order to more strongly identify with the picture. Then, a picture of the robot is shown to the child and placed next to the child's drawing. A collection of pictures is then shown to the child in random order, from which the child chooses the \emph{most appropriate} picture to place in between the robot and the drawing of themselves. Note that the interpretation of how to select the \emph{most appropriate} picture is left to the child. The researchers do not present the pictures in a predefined order, as not to bias their choice. 

\begin{figure}
\centerline{\includegraphics[width=3in]{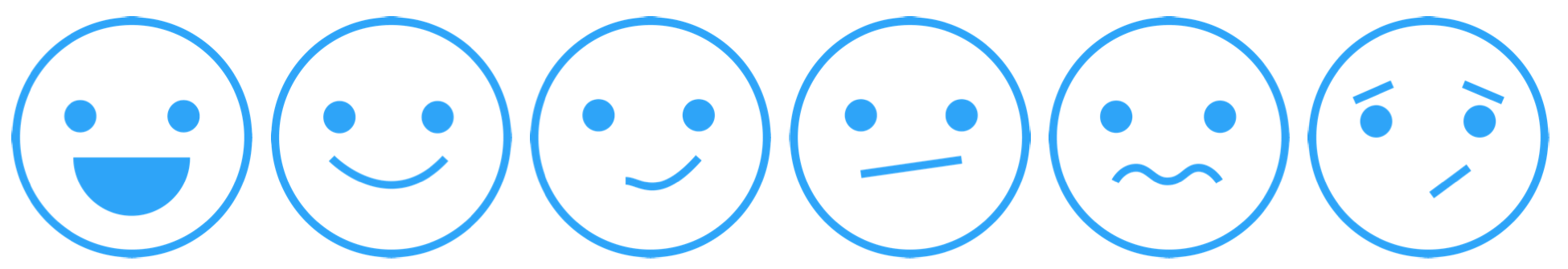}}
\caption{Collection of emotions shown to the child to identify the child's emotions towards the robot. The pictures are presented to the child as an unordered collection. The child is consecutively asked to choose and explain two pictures, which in his or her opinion best match the interaction with the robot.} \label{fig:emoticons}
\end{figure}

Similar to the Smileyometer described in the Fun Toolkit \cite{Read2002}, the picture collection consists of smiley faces displaying certain emotions, as shown in figure~\ref{fig:emoticons}. These pictures of emotions help the children describe their own emotions and feelings during the experiment and towards the robot. After choosing the most suitable picture, we ask the child to explain what emotion they see on the picture, and why this emotion fits between them and the robot. Subsequently, the child is asked to select and explain a \emph{second} picture, to allow us to explore possible nuances in the earlier answer: does the child incline more towards positive or negative emotions, or perhaps selects a combination of a positive and negative emotion?

\begin{figure}
\centerline{\includegraphics[width=3.5in]{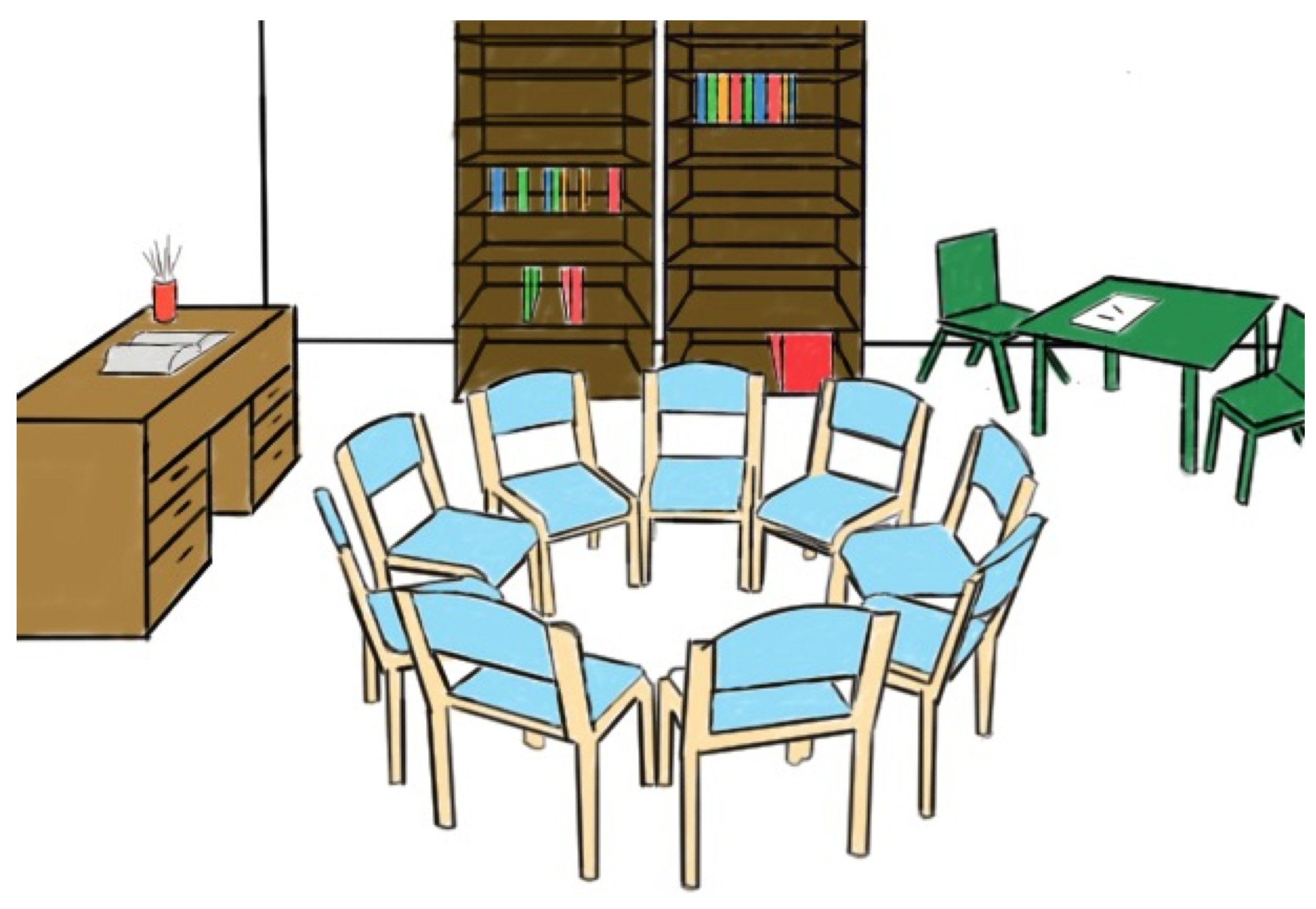}}
\caption{Classroom with circle of chairs, wherein the children are asked to place themselves, the robot and their other classmates.} \label{fig:classroom}
\end{figure}

Secondly, we attempt to measure the social distance between the child and the robot. A recognisable setting of a circle of chairs in a classroom is shown to the children, as illustrated in figure~\ref{fig:classroom}. In addition, the children received cartoon pictograms of the face of the robot and faces of various children. The child in the experiment chooses one of the available faces to be his or her own, while the other faces represent the classmates of the child. The child is then asked to place themselves, the robot and the other classmates into the room. Since kids generally like to sit next to their friends and socially close contacts, we assume this setup will give an indication on how socially close the child feels towards the robot. In addition, we ask the children to explain why they seat themselves and the robot in the specific locations.

Finally, we use questions adopted from the field of sociometry \cite{Busse2005}, supplemented with questions inspired by the work of Beran et al. \cite{Beran2011}. More specifically, we ask children whether they would invite the robot to their home, whether they would tell the robot a secret, whether they would share food with the robot, whether the robot could hear or see them, and whether they could be friends with the robot. 

The above methods are combined in the form of a semi-structured interview, to gain further insight in the child's reasoning and thoughts. The half-open questions focus on the child's opinion about the task, and their perceptions of the robot and themselves in the task. From these questions a matrix is drawn to describe the relation and engagement the child has with the robot and the task. For instance, a child may answer that he or she will tell the robot a secret, another child would say they don't have a secret, but would then invite Zeno to their home. Since all these questions refer to the surrounding of the child, it might be that one single answer is unsuitable to determine the nature of the relationship, while a combination of questions might lead to a more general insight on the relationship between robot and child. Additionally, we perform a qualitative analysis of the child's explanations, to gain more insight in the reasoning behind selecting certain answers.

Based on the collection of these three measurements, we compute an exploratory cumulative score. This score covers all the social relationship features that we have measured, and is composed of a sum of weighted scores for each individual measurement. We attempt to interpret this as a summary of each individual child's social relationship with the robot, without going into detail for each individual factor.

\section{PROCEDURE}
The overall setup of the experiment was the following: First, the researcher introduced the robot to the child. It was explained who the robot is and what would happen during the experiment. Then the learning task was explained to the child. Both groups, the intervention and the control group, got the same introduction to the task. After the introduction, the robot guided the child through the learning assignments. The assignments were additionally displayed on a tablet, to supply multiple channels of communication. The consecutive assignments in the inquiry task were of increasing difficulty, all children started with the same easy assignment and ended with the same difficult assignment. The first two assignments were the same in both conditions. The control group then received two additional assignments of intermediate difficulty, while the intervention group performed the physical extracurricular activity. The participants in both conditions then finished with the same two assignments. The semi-structured interview was run after the last assignment.

\subsection{Participants}
The experiments were conducted at a local daycare centre. Prior to participation, each child's parent or legal guardian was informed of the activities, study goals, and data collection methods, and was asked to fill out an informed consent form. After completing the experiments all children were given a central debriefing, where they could ask questions and say goodbye to the robot. This study was approved by the university's Electrical Engineering, Mathematics and Computer Science (EEMCS) ethics board.

Over the course of a week, a total of 23 children participated at the daycare centre. In the intervention condition the 12 participants (7 boys, 5 girls) had an average age of 8 years (SD = 2.73). In the control condition the 11 participants (9 boys, 2 girls) had an average age of 8.6 (SD = 1.65).

\section{RESULTS}
Based on our observations, generally the children's reactions on the experiment were very positive and enthusiastic. Overall, they enjoyed participating in the experiment and playing with the robot.

All children of the intervention group followed the robot's suggestion to stand next to him to perform the physical exercise. All of them followed the movements and suggestions of the robot. Even difficult tasks which the robot was unable to do himself were performed by the children. Only two of the tasks were misunderstood by a small amount of children. Some children misinterpreted the task ”drawing an eight in the air with your foot”. Instead of drawing an eight with their foot, they did it with their hand. Furthermore, we found that children did not continue to spell the alphabet after the first few letters. The children interpreted all other parts of the learning assignments and the intermediate physical exercise correctly.

The control group did like the task and the robot in general, but children indicated several times that the learning task was repetitive. None of the children of the intervention group said this about the learning tasks in combination with the physical exercise.

\subsection{Semi-structured interview}
Most of the children participated well in the interview and tried to answer the questions. Some were shy, which resulted in a less detailed interview with unanswered questions or missing explanations. The interview started with an open question about what they liked and disliked about the experiment. All children enjoyed the learning tasks, and some mentioned it was funny to do the experiments.

\paragraph{What did you think about the experiment and the robot?}
All the children answered that the assignment as well as the robot was nice. A few children answered more specifically that they liked the prediction step of the inquiry cycle, or that they liked one of the challenging learning tasks. At this point some children of the control group indicated that the learning task was always the same. Children of the intervention group often said, that they liked the physical exercise and overall the variety of assignments they did with the robot.

\paragraph{Pictorial emotion task}
Children were free to interpret the emotions, since the collection of pictures was presented in random order. Therefore, they were first asked which emotion they saw on the picture and why they chose this emotion. From these answers the child's interpreted emotions become apparent, as shown in table \ref{tab:emotion_interpretations}. Generally, the interpretations were very similar. However, a small minority interpreted the open-mouthed smiley as surprised or amazed.

\begin{table}
\caption{Children's interpretations of their picked emotions.} \label{tab:emotion_interpretations}
\centerline{\includegraphics[width=2.5in]{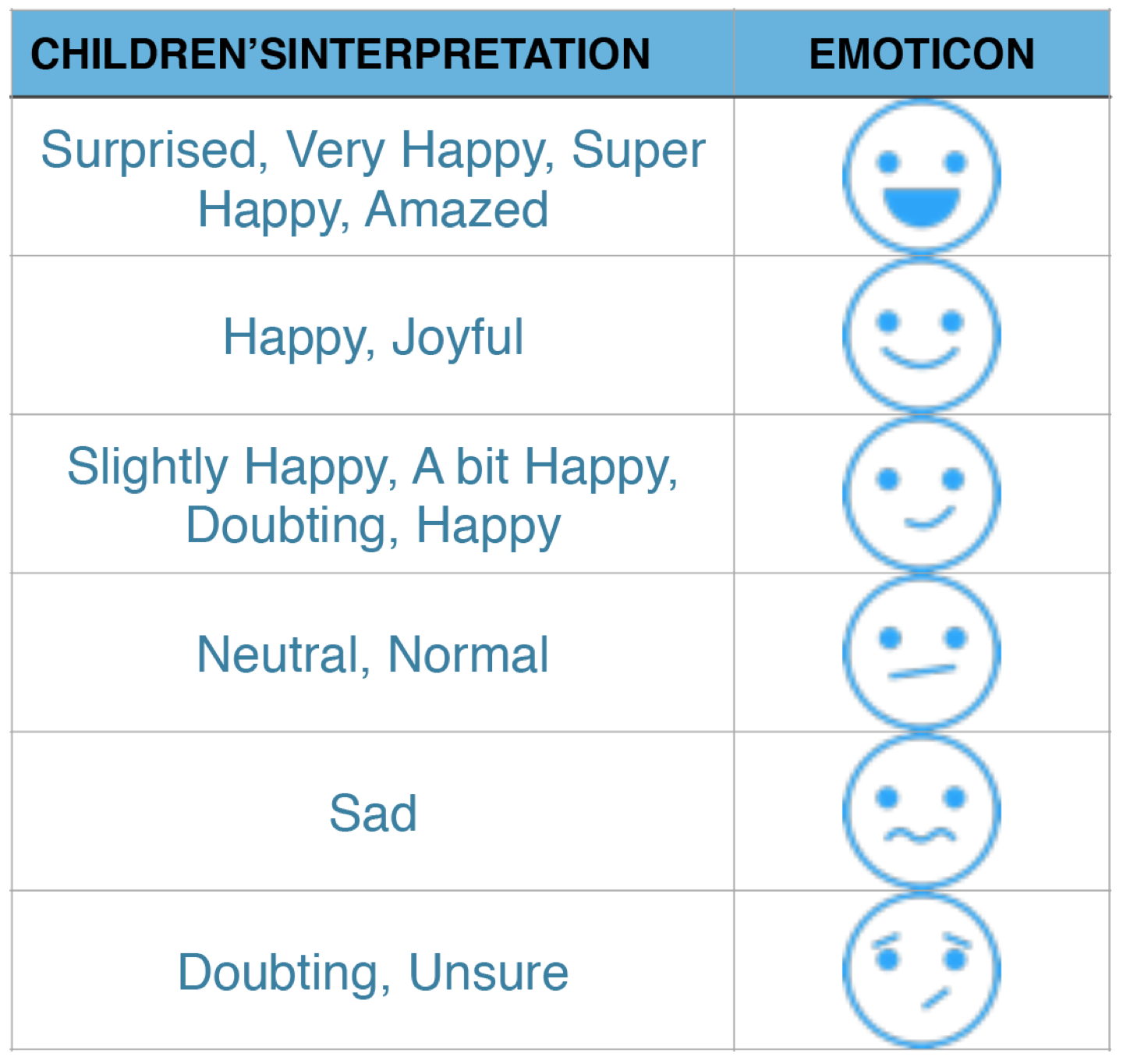}}
\end{table}

\begin{table}
\caption{Overview of children's first and second picked emotions. The combinations presented here show the amount of children who picked this specific combination of emotions for their first and second choice. Condition 1 is the intervention group, condition 2 is the control group.} \label{tab:emotion_combinations}
\centerline{\includegraphics[width=3.5in]{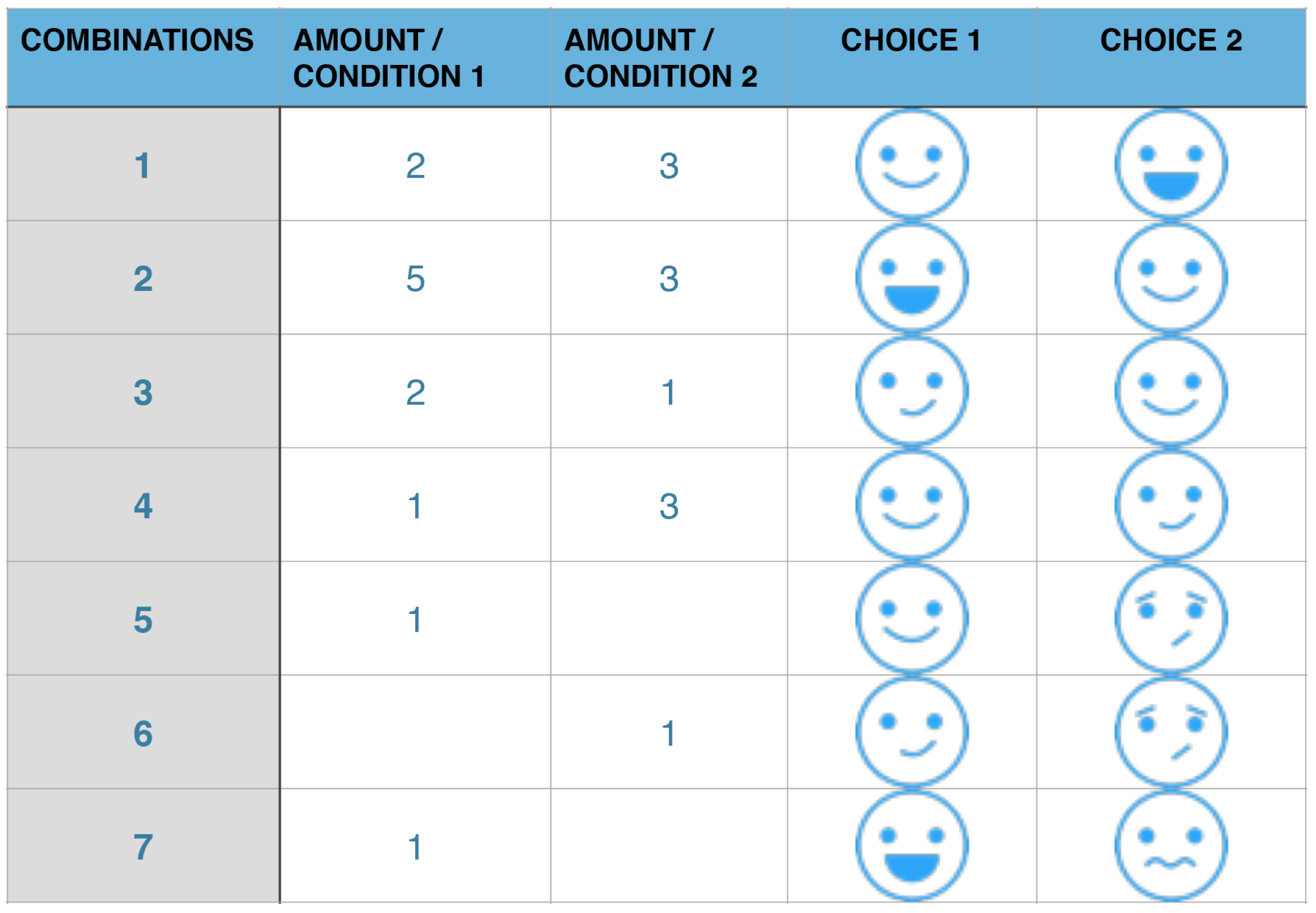}}
\end{table}

As described previously, the child would choose and explain two different pictures in succession, indicating which combination of two emotions best describe how they feel about their interaction with the robot. Table \ref{tab:emotion_combinations} shows that the children generally picked a combination of two happy emoticons to describe their emotions. In the intervention group 50\% of the children chose the ``happiest'' smiley as their first choice. For the control group, 27.3\% of the children chose the ``happiest'' smiley as a first choice. The two happy smileys have been used 83\% in the intervention group. The control group used these two smileys 81\%. Less happy smileys were selected 5 out of 24 for the intervention group and 6 times out of 22 for the control group.

The second combination in table \ref{tab:emotion_combinations}  has been used the most in the intervention group. This combination was selected by 41.6\% of the intervention group compared to 27.3\% of the control group. For the control group, combinations 1, 2 and 4 were each chosen in 27.3\% of the cases.

\paragraph{Social distance}
Most of the children put the robot directly next to themselves in the class circle. When asked why, they stated that they liked the robot or thought it was nice to sit next to him. Some projected a real relationship and indicated they would laugh with the robot or the robot could help them if he would sit next to them. A few children placed the robot somewhere else, stating that they wanted only their closest friends to sit next to them. In the intervention group, 83\% placed the robot directly next to themselves, compared to 90\% of the control group.

\paragraph{Friendship with robot}
The large majority of the children in both groups indicated that they could imagine a friendship with the robot: 83.3\% of the intervention group and 81.8\% of the control group considered a friendship with the robot. They reasoned with phrases like: ``Zeno can do everything'', ``He is very lovely'', ``We look like each other'' or ``We had fun together''. A low percentage of the children doubted before they answered and first asked, for instance, whether the robot could play soccer. From the intervention group 8.3\% and from the control group 9.1\% would not consider a friendship with the robot. In summary, children of the intervention group and control group showed a similar preference for becoming friends with the robot.

\paragraph{Age of the robot}
For the age of the robot, the children reasoned very differently. Some children observed the appearance of the robot in order to determine the age, some others included the introduced background of the robot, while others used their imagination. The majority of children in both conditions stated that the robot is slightly younger than themselves. A few children answered that the robot is either the same age or older. In average the age of the robot was guessed to be 8.6 years for the intervention group. In the control group the children mentioned an average age of 9.3 years.

\paragraph{Sharing a secret with the robot}
The children were asked whether they would tell the robot a secret. This question measures the trust the children would have in the robot \cite{Busse2005}. The children sometimes said they won't tell a secret because the robot is interacting with so many children, there is a high chance that the robot would tell it. Other children stated that they would only tell it, if the robot would not tell it to somebody else.

A slight majority of both groups stated they would tell the robot a secret (58.3\% in the intervention group, 54.5\% the control group). A smaller percentage would not want to tell the robot a secret (41.67\% in the intervention group, 27.3\% the control group). Of the control group, 18.2\% did not respond to the question. Most of the children who would not tell the robot a secret, still said that the robot could be a friend. 

\paragraph{Invitation to the child's home}
When we asked the children whether they would invite the robot to their home, some of the children got very excited about this idea. Of the intervention group, 91.6\% wanted to invite the robot to their home, while 8.3\% did not respond to the question. From the control group, 63.6\% would invite the robot to their home, 18.2\% of the children answered they would not invite the robot to their home and 18.2\% did not respond to the question.

As a follow up question, we asked what the child and the robot would do at home. The answers to this question were very broad. Of the intervention group, 75\% proposed some sort of activity, all of which unrelated to the experiment, while 25\% did not respond to the question. The control group proposed 36.3\% related activities and 18.1\% unrelated activities, while 45.5\% did not respond. Clearly, the children that experienced the physical activity together with the robot were able to imagine a broader range of activities they could share with the robot. Nonetheless the imagination of the children in general was very broad.

\paragraph{Sharing food with the robot}
The children were asked whether they would be willing to share their food with the robot. This question was intended to trigger the child's sociality towards the robot. All children of the intervention group wanted to share food with the robot. Of the control group, 27.3\% rejected to share their food with the robot.

\paragraph{Audition and sight of the robot}
When asked about whether or not the robot could see and hear, both groups were equally confused on whether the robot has sight, audition or both. Some children indicated, that it was one of the researchers, who controlled the robot. Others thought the tablet or other technology enables the system to function autonomously.

In the intervention group, 50\% of the children indicated that the robot has sight and 66\% stated that it could hear. Two children stated that the robot has none of the two. In the control group, 45.5\% of the children said that the robot can see and 54\% said that it could hear. Three of the asked children stated that the robot has neither sight nor audition.

\subsection{Descriptions of the robot}
During the interview, most children commented on the robot's appearance or behaviour. Many children thought the robot was funny. They commented that the robot had a female voice and a male appearance. The robot was often described as nice, smart and happy. Some children thought he was not that happy or that they expected him to  be taller. Often, the children were surprised that he could talk and perform various complex body movements. Most of the children thought the robot was slightly younger than themselves, however a few kids thought he was an adult and looked older than them.

The role of the robot often appeared in the interview to be very diverse. Some children commented that \emph{they would help the robot} understand things, while others said \emph{the robot could help them} in studying. One child answered that the robot could help clean his room. Another child liked that the robot was neither a teacher nor a peer student, but that it was nice that the robot was of ``a different kind''.

During the interview the children would often indicate some descriptive characteristics of the robot. For example, some described the robot the same as themselves, similar to them, or that he could do everything. Some indicated that his movements were quite natural or that they thought that robots could do less. On the other hand, some said the robot could not do all of what they themselves could do.

\subsection{Cumulative analysis}
Due to the fact that each individual measure only looks at one very specific relationship aspect, we propose an additional \emph{exploratory cumulative analysis}. For this analysis we calculate a cumulative ``social relationship'' score for each child, based on the individual answers and measurements mentioned above. The expectation is that this cumulative score contains some information about the general perceived social relationship they have with the robot, although we will not be able to identify the individual aspects which influence this relationship.

The assigned measures included most of the questions the child answered. All these questions receive a certain score, depending on how the child answered the question. The different scores for each metric can be found in table \ref{tab:cumulative_scores}. Friendship gets a relatively high score, due to the fact that considering friendship is a very important criteria for building a social relationship. Additionally, the other scores are chosen in such a way that measures with a high variance get a higher value than the measures with a low variance. Questions such as the age are left out, as the child's reasoning behind this has been very different: the children's answers did not always represent a perception of the robot's age. 

 \begin{table}
\caption{Scores applied for various measures, as used in the exploratory cumulative analysis. For each sociometric measure, a score of 0, 5, 10 or 15 is added to the child's cumulative score if answered ``yes'', or 0 if the child answered ``no''. For the pictorial emotions task, a score of 1, 3, 5 or 10 is added, depending on the selected emotion for their first and second choice.} \label{tab:cumulative_scores}
\centerline{\includegraphics[width=3in]{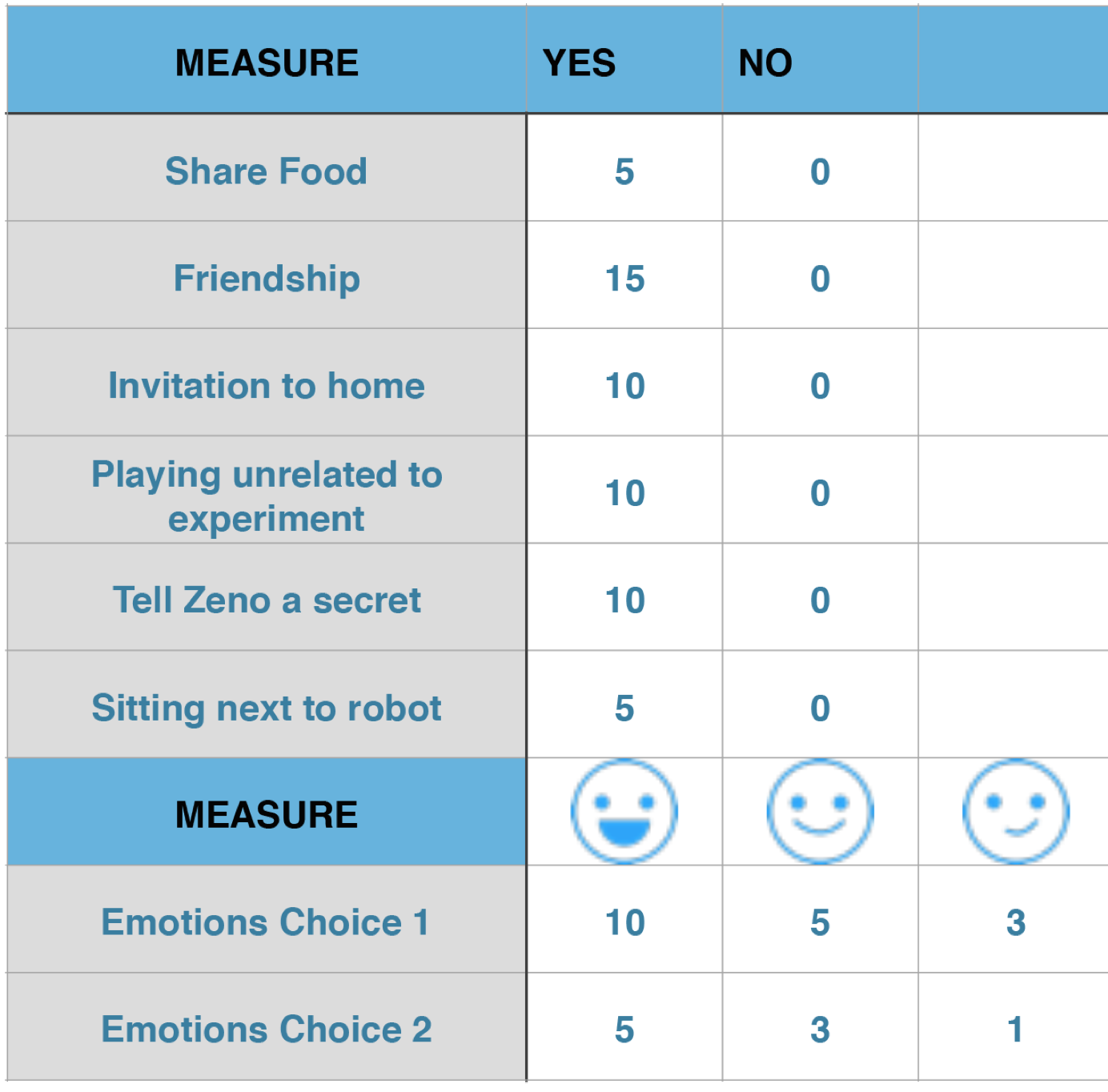}}
\end{table}

\begin{figure}
\centerline{\includegraphics[width=3.5in]{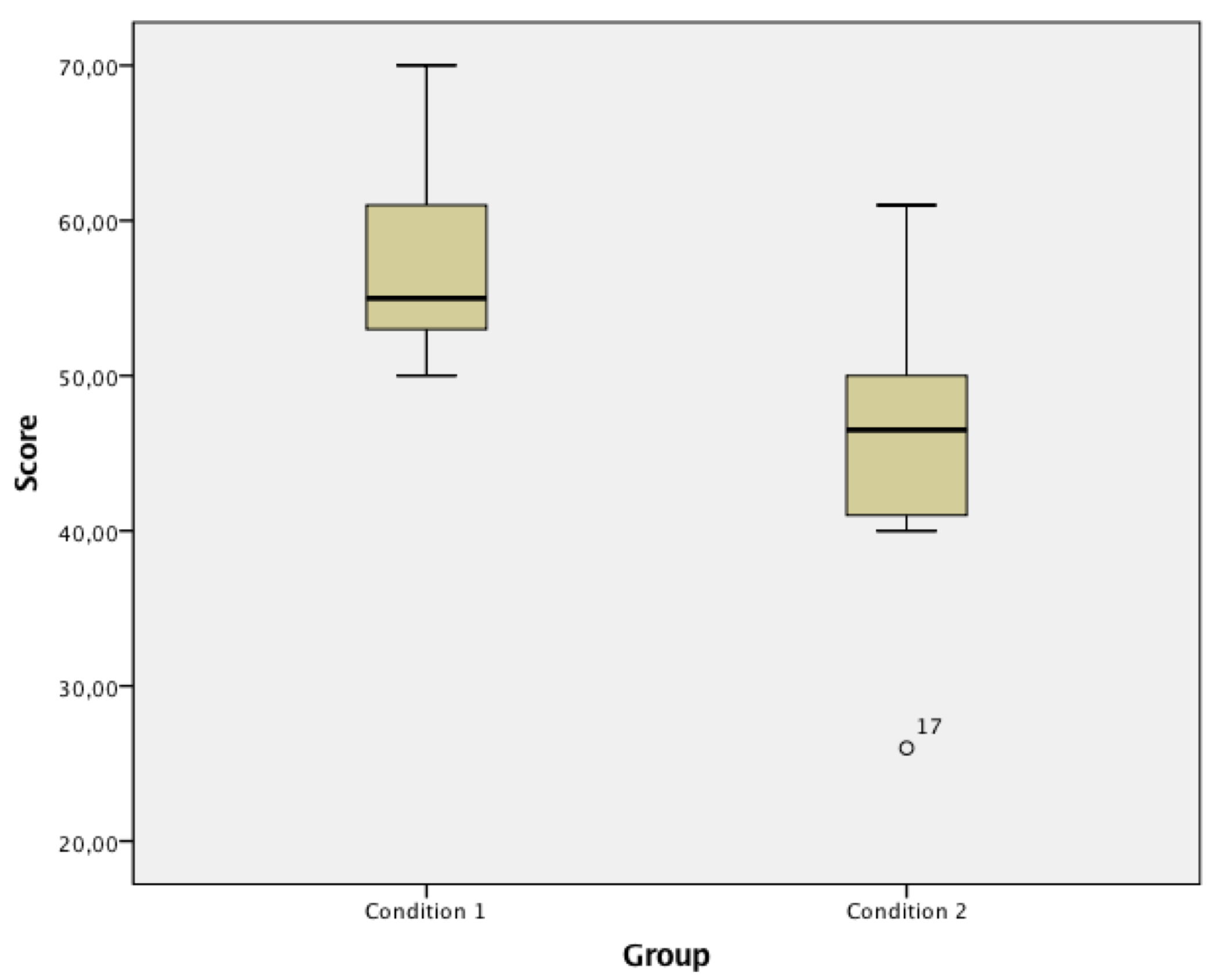}}
\caption{Box plot showing the distribution of the exploratory cumulative scores, for the intervention group (condition 1) and the control group (condition 2).} \label{fig:boxplot_cumulative_means}
\end{figure}

Results from two children were excluded from this analysis, due to their high amount of unanswered questions. The mean of the intervention group is 57.5 (n = 11, SD = 7.19) and the mean of the control group is 45.7 (n = 10, SD = 9.59). A Wilcoxon-Mann-Whitney test shows a significant (U = 15, p \textless 0.005) difference between the intervention condition and the control group. Figure \ref{fig:boxplot_cumulative_means} shows the distribution in scores between both conditions.

\section{DISCUSSION}
From observations we get the impression that most of the children enjoyed the overall experiment and specifically the physical exercise with the robot. However, a more detailed analysis of the recorded sessions is needed to further investigate this. So far, we base this subjective impression on the preliminary assessment of the videos and the reactions that came from the childcare supervisors and the children themselves. Nonetheless, this is worth mentioning, since an enjoyable and nice task is fundamental to keeping children engaged with learning.

\subsection{Interview}
We interpret the individual interview results as some of the parameters that describe a relationship. Although we recognise that we only touched upon a fraction of the factors influencing something as complex as social relationship building, we tried to research aspects of trust, sociality, friendship, social distance and the children's general view and perception of the robot. It can be concluded that several of the questions did not show any difference between the two conditions. This means that there is no clear indication that the physical extracurricular activity influences the specific factors of the relationship that we asked about. 

A reason for the similar answers of the children in both conditions could be that the questions were not asked in an operative understandable way. This is always a challenge with children. These rather inoperative questions were asking about ``sharing food with the robot'' and the ``social distance'' (the chair seating assignment). These two measures were answered in the same way by almost all children. Other measures such as ``an invitation to their home'' or ``considering a friendship with the robot'' showed more variance. However, many children still answered in the same way, which lead to an unclear view on these measures. 

Results from the interviews suggest that for some questions there is a slight difference between the two conditions, even though this difference might not be very strong. In questions about ``inviting the robot to the home'', ``the type of playing at home'', and ``sharing food'', indications can be found that the intervention group had a different view of the robot, due to their more varied experience with the robot. This might have influenced their imaginations of what else the robot is capable of. The question whether the children would tell a secret to the robot, resulted in a mixed outcome. In terms of numbers, from the intervention group 58.3\% would tell Zeno a secret. From the control group, 54.5\%would tell a secret. Thus, the difference is again very small.

Similar results occurred regarding the consideration of a friendship with the robot. The difference between the two groups was very small, which complicates drawing any conclusion. Therefore, it seems that most of the children would consider a friendship with Zeno despite of the type of interaction they had with the robot.

The emotion assignment showed some differences, however these are difficult to interpret. The first picture they chose often described their general feeling, the second picture seemed to identify whether they clearly tend into a certain direction in terms of emotions. For example, if the first picture was a normal smiling/happy emotion, then the second choice could identify the direction and verify the choice (either very happy, or moderately happy). 

This was however not always the case. Sometimes, the child expressed two different emotions that he or she could relate to the robot. For instance one child also chose a sad emoticon, stating that he could also come to talk to the robot when he was feeling sad. A few children interpreted the picture with the broad smile as being surprised. The slightly smiling emoticon has often been interpreted as a ``little bit happy'', but also sometimes as ``doubting''. 

However, it can be seen that there are more ``happy'' emotions selected by the intervention group. This shows that there is a possibility that, overall, the intervention group indeed felt happier about their interaction with the robot. The combination showing the two happiest smileys, while choosing the happiest smiley as a first choice, was chosen 41.6\% for the intervention group, compared to 27.3\% for the control group.

The assignment where the children put themselves, their friends and the robot into the classroom, did not result in an indication of social distance. The robot was placed next to the child in almost all cases, revealing no clear difference between the conditions.

Although there are some non-significant indications in favour of the intervention condition for some of the measures, this is not reflected in all measurements. Whether there has been an actual difference in the relationship is difficult to determine, due to the fact that some measures showed unclear results and in general it is difficult to measure relationships, especially for such a short interaction. Hence, a more robust per case interpretation is needed.

\subsection{Cumulative analysis}
The exploratory cumulative analysis shows a difference between the two conditions. This analysis should be interpreted with care, though, since the weights for each measure were determined intuitively. This analysis gives an estimated indication of the social relationship between the child and the robot, since all respective relationship measures are summarised and taken into account per child. 

We have shown that the scores for the intervention group are significantly higher than for the control group. However, due to the exploratory nature of this cumulative score, we are unable to specify exactly which factors have resulted in the difference found between the two conditions. 

\subsection{Limitations}
Partly due to the exploratry nature of this study, we discuss several limitations that should be taken into account when interpreting the results. 

Since the child engages in an activity with the robot that is both \emph{physical} and \emph{extracurricular} in nature, any measured effects could be attributed to one of the following: 1) the physical nature of the activity; 2) the extracurricular nature of the activity; or 3) a combination of both. Repeating the experiment with an additional condition containing a ``passive'' extracurricular activity would allow us to explore this effect in more detail.

Some children reported the control condition as being ``boring''. Structured inquiry tasks are repetitive by nature, and although the consecutive tasks were selected in such a way that there was a steady increase in difficulty, this repetitive effect seemed more pronounced in the control condition. It is therefore unclear if the children's answers are a reflection of their level of enjoyment, or their perceived relationship with the robot. A repeated experiment could investigate this in more detail, where all children engage in identical learning tasks and extracurricular activities. In two conditions, they would either to the extra activity alone or in collaboration with a social robot.

The methods used in the semi-structured interview often show very similar results between conditions, making it difficult to interpret the effects of the manipulation. Only in the cumulative analysis do we see a difference emerging between the control and intervention group. More research is needed to validate the methods and determine underlying constructs, which will influence the weights of the cumulative function.

\section{CONCLUSIONS}
In the study presented here, a child and a robot work together on a structured inquiry learning task. We investigate whether the child's perceived relationship with the robot is influenced by engaging in a shared physical extracurricular activity with the robot. Measurements are gathered using a semi-structured interview, which is composed of a pictorial task, a social distance task, and several sociometric questions. Generally, the children seemed to enjoy working with the robot, indicating that they would invite him home, or that they could become friends. Results for most individual measurements are inconclusive, however. The pictorial taks, where children pick emotion cards that fit their relationship with the robot, seemed to give promising results: most children gave similar descriptions of the depicted emotions, and generally picked more positive emotions in the condition where they engaged in the shared activity with the robot. Finally, the cumulative score that aggregates all used measures into a single value revealed a difference between conditions, although the limitations make it difficult to further interpret this result.

Since the first insights from this study seem promising, future work will focus on further exploration and verification of the cumulative analysis method proposed in this paper, as well as the individual measurement methods used during the semi-structured interview. Additionally, we aim to investigate how a change in relationship between child and robot impacts the child's interactions with the learning materials, and consequently the child's learning methods and learning performance.

\ack
This project has received funding from the European Union Seventh Framework Programme (FP7-ICT-2013-10) as part of EASEL under grant agreement no 611971.

\bibliographystyle{AISB2008}
\balance
\bibliography{AISB2016_arXiv_version.bbl}

\begin{thebibliography}{10}

\bibitem{Belpaeme2012}
Tony Belpaeme, Paul~E Baxter, Robin Read, Rachel Wood, Heriberto
  Cuay{\'{a}}huitl, Bernd Kiefer, Stefania Racioppa, Ivana
  Kruijff-Korbayov{\'{a}}, Georgios Athanasopoulos, Valentin Enescu, Rosemarijn
  Looije, Mark Neerincx, Yiannis Demiris, Raquel Ros-Espinoza, Aryel Beck, Lola
  Ca{\~{n}}amero, Antione Hiolle, Matthew Lewis, Ilaria Baroni, Marco Nalin,
  Piero Cosi, Giulio Paci, Fabio Tesser, Giacomo Sommavilla, and Remi Humbert,
  `{Multimodal Child-Robot Interaction: Building Social Bonds}', {\em Journal
  of Human-Robot Interaction}, {\bf 1}(2),  33--53, (jan 2013).

\bibitem{Beran2011}
Tanya~N. Beran, Alejandro Ramirez-Serrano, Roman Kuzyk, Meghann Fior, and Sarah
  Nugent, `{Understanding how children understand robots: Perceived animism in
  child–robot interaction}', {\em International Journal of Human-Computer
  Studies}, {\bf 69}(7-8),  539--550, (jul 2011).

\bibitem{Busse2005}
R~Busse, `{Sociometric Assessment}', in {\em Encyclopedia of school
  psychology}, ed., S~Lee,  520--521, SAGE Publications, Inc., Thousand Oaks,
  CA, (2005).

\bibitem{Chandra2015}
Shruti Chandra, Patricia Alves-Oliveira, Severin Lemaignan, Pedro Sequeira, Ana
  Paiva, and Pierre Dillenbourg, `{Can a child feel responsible for another in
  the presence of a robot in a collaborative learning activity?}', in {\em 2015
  24th IEEE International Symposium on Robot and Human Interactive
  Communication (RO-MAN)}, pp. 167--172. IEEE, (aug 2015).

\bibitem{Charisi2015}
Vicky Charisi, Daniel Davison, Frances Wijnen, Jan {Van Der Meij}, Dennis
  Reidsma, Tony Prescott, Wouter {Van Joolingen}, and Vanessa Evers, `{Towards
  a child-robot symbiotic co-development : A theoretical approach}', in {\em
  New Frontiers in Human-Robot Interaction}. AISB, (2015).

\bibitem{Harter1984}
S~Harter, R~Pike, and Child Development, `{The pictorial scale of perceived
  competence and social acceptance for young children}', {\em Child
  development}, {\bf 55}(6),  1969--1982, (1984).

\bibitem{Read2002}
Stuart~Macfarlane {Janet Read}, `{Endurability, Engagement and Expectations:
  Measuring Children's Fun}', in {\em Interaction Design and Children}, pp.
  1----23. Shaker Publishing, (2002).

\bibitem{Kahn2012}
Peter~H Kahn, Takayuki Kanda, Hiroshi Ishiguro, Nathan~G. Freier, Rachel~L.
  Severson, Brian~T. Gill, Jolina~H. Ruckert, Solace Shen, and Peter~H. {Kahn
  Jr.}, `{"Robovie, you'll have to go into the closet now": children's social
  and moral relationships with a humanoid robot.}', {\em Developmental
  psychology}, {\bf 48}(2),  303--14, (mar 2012).

\bibitem{Kanda2007}
T.~Kanda, R.~Sato, N.~Saiwaki, and H.~Ishiguro, `{A Two-Month Field Trial in an
  Elementary School for Long-Term Human–Robot Interaction}', {\em IEEE
  Transactions on Robotics}, {\bf 23}(5),  962--971, (oct 2007).

\bibitem{Kanda2012}
Takayuki Kanda, Michihiro Shimada, and Satoshi Koizumi, `{Children learning
  with a social robot}', in {\em Proceedings of the seventh annual ACM/IEEE
  international conference on Human-Robot Interaction - HRI '12}, p. 351, New
  York, New York, USA, (mar 2012). ACM Press.

\bibitem{Klahr2000}
David Klahr, {\em {Exploring Science: The Cognition and Development of
  Discovery Processes}}, The MIT Press, Cambridge, 2000.

\bibitem{Klahr1988}
David Klahr and Kevin Dunbar, `{Dual Space Search During Scientific
  Reasoning}', {\em Cognitive Science}, {\bf 12}(1),  1--48, (jan 1988).

\bibitem{Kory2014}
Jacqueline Kory and Cynthia Breazeal, `{Storytelling with robots: Learning
  companions for preschool children's language development}', in {\em The 23rd
  IEEE International Symposium on Robot and Human Interactive Communication},
  pp. 643--648. IEEE, (aug 2014).

\bibitem{Read2006}
Janet~C. Read and Stuart MacFarlane, `{Using the fun toolkit and other survey
  methods to gather opinions in child computer interaction}', in {\em
  Proceeding of the 2006 conference on Interaction design and children - IDC
  '06}, p.~81, New York, New York, USA, (jun 2006). ACM Press.

\bibitem{Rogoff1998}
Barbara Rogoff, `{Cognition as a collaborative process}', in {\em Handbook of
  child psychology}, chapter Cognition,,  679--744, John Wiley {\&} Sons Inc,
  Hoboken, NJ, US, (1998).

\bibitem{Joolingen1997}
Wouter~R. van Joolingen and Ton de~Jong, `{An extended dual search space model
  of scientific discovery learning}', {\em Instructional Science}, {\bf 25}(5),
   307--346, (1997).

\bibitem{Vygotsky1978}
L.S. Vygotsky, {\em {Mind in Society: The Development of Higher Psychological
  Processes}}, Harvard University Press, Cambridge, MA, 1978.

\bibitem{Wijnen2015}
Frances Wijnen, Vicky Charisi, D.P.~Daniel Davison, J.~van~der Meij, Dennis
  Reidsma, and Vanessa Evers, `{Inquiry learning with a social robot: can you
  explain that to me?}', in {\em New Friends 2015: the 1st international
  conference on social robotics in therapy and education}. Windesheim
  Flevoland, (oct 2015).

\bibitem{Zimmerman2000}
Corinne Zimmerman, `{The Development of Scientific Reasoning Skills}', {\em
  Developmental Review}, {\bf 20}(1),  99--149, (mar 2000).

\end{thebibliography}

\end{document}